\documentclass[10pt,twocolumn,letterpaper]{article}

\usepackage[pagenumbers]{cvpr} 

\makeatletter
\@namedef{ver@everyshi.sty}{}
\makeatother

\usepackage{amsmath,amsfonts,amsthm,amssymb}
\usepackage{mathtools}
\usepackage{bm}
\usepackage{nicefrac}
\usepackage{microtype}

\usepackage{color,xcolor}
\usepackage{epsfig}
\usepackage{graphicx}

\usepackage{adjustbox}
\usepackage{array}
\usepackage{booktabs}
\usepackage{colortbl}
\usepackage{wrapfig}
\usepackage{hhline}
\usepackage{multirow}
\usepackage{subcaption}
\usepackage[size=small]{caption}

\usepackage{changepage}
\usepackage{extramarks}
\usepackage{fancyhdr}
\usepackage{lastpage}
\usepackage{setspace}
\usepackage{soul}
\usepackage{xspace}

\usepackage{url}

\usepackage{algpseudocode}
\usepackage{algorithmicx}
\usepackage[ruled]{algorithm2e}
\usepackage{enumerate}
\usepackage{todonotes} 
\usepackage{enumitem}  
\usepackage{makecell}
\usepackage{pifont}

\newcolumntype{L}[1]{>{\raggedright\let\newline\\\arraybackslash\hspace{0pt}}m{#1}}
\newcolumntype{C}[1]{>{\centering\let\newline\\\arraybackslash\hspace{0pt}}m{#1}}
\newcolumntype{R}[1]{>{\raggedleft\let\newline\\\arraybackslash\hspace{0pt}}m{#1}}

\newcommand{\ignore}[1]{}

\DeclareMathAlphabet{\mathbfit}{OML}{cmm}{b}{it}

\makeatletter
\DeclareRobustCommand\onedot{\futurelet\@let@token\@onedot}
\def\@onedot{\ifx\@let@token.\else.\null\fi\xspace}

\def\etal{et al\onedot}

\makeatother

\definecolor{MyDarkBlue}{rgb}{0,0.08,1}
\definecolor{MyAqua}{rgb}{0,0.7,0.7}
\definecolor{MyDarkGreen}{rgb}{0.02,0.6,0.02}
\definecolor{MyDarkRed}{rgb}{0.8,0.02,0.02}
\definecolor{MyDarkOrange}{rgb}{0.40,0.2,0.02}
\definecolor{MyPurple}{RGB}{111,0,255}
\definecolor{MyRed}{rgb}{1.0,0.0,0.0}
\definecolor{MyGold}{rgb}{0.75,0.6,0.12}
\definecolor{MyDarkgray}{rgb}{0.66, 0.66, 0.66}

\SetKwFor{For}{for }{}{}

\newcommand{\modelfull}{Equivariant Object detection Network\xspace}
\newcommand{\model}{EON\xspace}
\newcommand{\ourvn}{EON-VoteNet\xspace}

\def\bR{\mathbb{R}}

\newcommand{\myparagraph}[1]{\vspace{0.05cm}\noindent\textbf{#1}}

\usepackage[pagebackref,breaklinks,colorlinks]{hyperref}

\usepackage[capitalize]{cleveref}
\crefname{section}{Sec.}{Secs.}
\Crefname{section}{Section}{Sections}
\Crefname{table}{Table}{Tables}
\crefname{table}{Tab.}{Tabs.}

\def\cvprPaperID{1958} 
\def\confName{CVPR}
\def\confYear{2022}

\begin{document}

\title{Rotationally Equivariant 3D Object Detection}

\author{Hong-Xing Yu\\
Stanford University\\
\and
Jiajun Wu\\
Stanford University\\
\and
Li Yi\\
Tsinghua University, Shanghai Qi Zhi Institute\\
}

\maketitle

\begin{abstract}
    Rotation equivariance has recently become a strongly desired property in the 3D	deep learning community. Yet most existing methods focus on equivariance regarding a global input rotation while ignoring the fact that rotation symmetry has its own spatial support. Specifically, we consider the object detection problem in 3D scenes, where an object bounding box should be equivariant regarding the object pose, independent of the scene motion. This suggests a new desired property we call object-level rotation equivariance.
    To incorporate object-level rotation equivariance into 3D object detectors, we need a mechanism to extract equivariant features with local object-level spatial support while being able to model cross-object context information. To this end, we propose \modelfull (\model) with a rotation equivariance suspension design to achieve object-level equivariance. \model can be applied to modern point cloud object detectors, such as VoteNet and PointRCNN, enabling them to exploit object rotation symmetry in scene-scale inputs.
    Our experiments on both indoor scene and autonomous driving datasets show that significant improvements are obtained by plugging our \model design into existing state-of-the-art 3D object detectors. Project website: \url{https://kovenyu.com/EON/}.
\end{abstract}
\section{Introduction}





3D object detection is a fundamental problem in various downstream applications including augmented reality, robotics, and autonomous driving. 
Research efforts in designing 3D object detection networks have shown great effectiveness for both indoor \cite{qi2019deep,xie2020mlcvnet} and outdoor scenes \cite{yin2021center,shi2019pointrcnn}. 
However, existing 3D object detectors cannot explicitly treat object rotation equivariance in their designs. Object rotation equivariance in 3D detection is well-reflected in the rotation invariance of shape and equivariance of orientation. 
For example, no matter how an object is oriented in an input scene, the detection result (typically represented as an oriented bounding box) associated with the object should orient in the same way while retaining the same shape.
Explicit modeling of these strong priors save the needs for expensive data augmentations, and can increase the expressivity and discriminative power of detection models without heavily increasing the number of parameters and introducing additional optimization challenges.

A recent trend to explicitly exploit rotation equivariance is through equivariant networks \cite{cohen2016group,worrall2017harmonic,thomas2018tensor} (EN). The main idea is that the equivariant geometric features carry both shape information and orientation information separately by design. Rotation equivariant networks have been explored for object classification and pose estimation \cite{deng2021vector,chen2021equivariant,li2021leveraging}, but not yet for 3D object detection. A main challenge is that existing EN mostly explores rotation equivariance regarding the full visual input, while equivariance to rotations of a whole scene is not ideal for object detection, because individual object orientation can be independent of the scene.
Thus, it is unclear how to achieve object-level rotation equivariance and how to benefit 3D object detection in cluttered scenes.

\begin{figure}[t]
    \centering
    \includegraphics[width=1.0\linewidth]{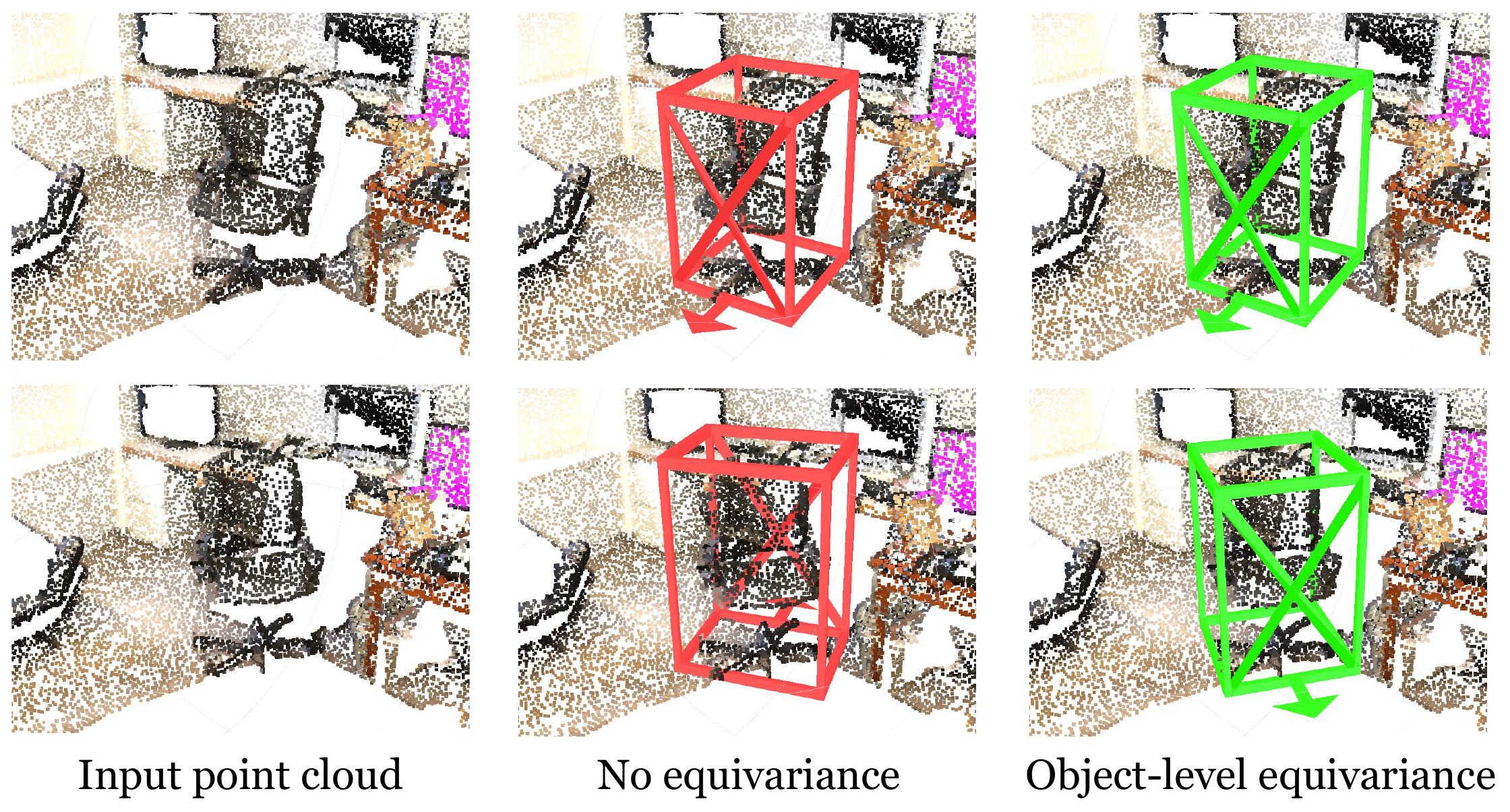}
    \caption{Rotation symmetry in object detection includes the equivariant orientation and invariant shape of the bounding box. The object-level rotation equivariant detector explicitly models these strong priors and intrinsically generates oriented bounding box to rotate following the object, while maintaining the same box shape. Non-equivariant detectors, however, may suffer from box shape changes and unaligned variations in box orientation.}
    \vspace{-0.3cm}
    \label{fig:teaser}
\end{figure}

We identify two key technical challenges toward object-level equivariant model design: how to determine the object-level spatial support to extract locally equivariant features, and how to aggregate context information. The greater context (such as nearby objects) is helpful to recognize objects especially in noisy or incomplete raw point clouds. However, the context information could easily break object-level rotation equivariance if not handled properly. For example, when detecting a chair, the presence of a nearby desk can provide useful context due to high co-occurrence probability. However, in case that the chair remains static while the desk changes its orientation, the chair features could also be affected unexpectedly.

We propose \modelfull (\model) to exploit object-level equivariance for 3D detection. Our core design is called \textbf{rotation equivariance suspension}. To properly determine the object-level spatial support, we let our model extract equivariant features only up to an intermediate stage. This is based upon the observation that most 3D detection networks extract features in a hierarchical manner \cite{qi2019deep,shi2019pointrcnn,liu2021} where early stages focus on local features while later stages cover more context-level information. Computing equivariant features only up to an intermediate stage produces local spatial support to rotation equivariance, and the model can adaptively learn to adjust its effective spatial support \cite{luo2016understanding}.
To allow aggregating context information, we \emph{suspend} equivariant feature computation at the previous intermediate stage by decomposing each equivariant feature into an object orientation hypothesis (orientation information) and an invariant object-frame feature (shape information). Our model keeps aggregating object-frame features in the latter stages, and finally resumes the orientation information for object proposals.  Since only object-frame features (i.e., without object orientation information) are aggregated after the intermediate stage, the greater contexts can be modeled without breaking object-level equivariance.

Our approach follows the modular design adopted by most bottom-up detectors, so that it can be easily plugged into state-of-the-art 3D object detectors. We have tested our method using various backbones and models on both indoor and outdoor 3D object detection benchmarks. We find that \model significantly boosts the performance of previous state-of-the-art 3D object detectors (+9.0 mAP on ScanNetV2, +3.1 mAP on SUN RGB-D, and +1.4 mAP on KITTI Dataset). In summary, our contributions are threefold:
\begin{itemize}
    \item To our best knowledge, this is the first work to explore rotation equivariance for 3D object detection.
    \item We propose \modelfull (\model), incorporating a novel design dubbed rotation equivariance suspension to exploit object-level equivariance in 3D detection. Our \model can be easily plugged into state-of-the-art bottom-up 3D object detectors.
    \item On both indoor and outdoor datasets, we demonstrate the benefits of object-level equivariance by boosting performances of previous state-of-the-art 3D detectors. 
\end{itemize}
\section{Related Work}

\myparagraph{3D object detection.} Most current state-of-the-art approaches to 3D object detection directly takes 3D data (such as point cloud, Lidar, and voxelized grids from them) as input and generates 3D oriented bounding boxes (OBBs) to represent the objects \cite{zhu2019class,zhang2020h3dnet,shi2020point,shi2020pv,zhou2018voxelnet,pan20213d,lang2019pointpillars,qi2019deep,yin2021center,shi2019pointrcnn}. Most of them follow a bottom-up design, where a backbone network extracts a sparse set of regional features from the dense input data, and a detection head proposes candidate OBBs (one-stage) or regions of interests (two-stage) for further refinement. Seminal works include VoteNet \cite{qi2019deep} and its follow-ups. VoteNet's backbone design includes two PointNet++, with a voting in between to help the aggregation of object surface points. H3DNet \cite{zhang2020h3dnet} improves VoteNet predictions using 3D primitives and a geometric loss. MLCVNet \cite{xie2020mlcvnet} further allows VoteNet to aggregate global contextual information via self-attention. Another line of methods is directly inspired from 2D bottom-up detectors, such as PointRCNN \cite{shi2019pointrcnn}, VoxelNet \cite{zhou2018voxelnet}, PointPillar \cite{lang2019pointpillars} and CenterPoint \cite{yin2021center}. However, these existing methods do not explicitly exploit object rotation equivariance in their models. Our method is grounded on this popular bottom-up modular design, and we focus on equipping these state-of-the-art detection models with object-level equivariance.

\myparagraph{Rotation equivariance networks.} 
From the seminal work of Group Equivariant Convolutional Network \cite{cohen2016group}, leveraging group equivariance for deep networks becomes increasingly popular. Existing approaches toward rotation equivariance can be roughly divided into two categories: filter orbit-based and filter design-based. Deriving from the group equivariant convolution \cite{cohen2016group,cohen2019gauge}, filter orbit-based methods discretize the rotation group and construct a set of group-transformed kernels (an ``orbit'') for the group equivariant computation \cite{chen2021equivariant,li2021leveraging,li2019discrete}. Filter design-based methods design intrinsically rotation-equivariant basis functions (e.g., generalized Fourier basis functions \cite{thomas2018tensor,cohen2018spherical}) and compose their networks with these basis functions \cite{thomas2018tensor,deng2021vector,cohen2018spherical,esteves2018learning}. Our method is inspired from the filter orbit-based Equivariant Point Network \cite{chen2021equivariant} (EPN) which introduces a tractable approximation to SE(3) group equivariant convolution on point clouds. However, EPN focuses on single object tasks and achieves equivariance regarding the full visual content. We target object-level equivariance for 3D detection in scenes.
Recently, there are a few works attempting to leverage rotation equivariance for aerial image detection \cite{han2021redet,yang2021sampling}. While they focus on 2D images, we aim at 3D object-level equivariant detection.
\section{Preliminary}

 
To compute rotation equivariant features, our approach finds inspiration from a recent state-of-the-art equivariant network, Equivariant Point Network (EPN)~\cite{chen2021equivariant}, that is designed for single objects. We briefly review EPN and provide an intuitive explanation for the equivariance property.


The key idea of EPN to achieve rotation equivariance is to augment each point feature vector $\mathbf{f}\in\bR^C$ to an equivariant feature $\mathbf{f}_\text{eqv}\in\bR^{C\times |G|}$ w.r.t. a discretized SO(3) subgroup $G$. Each vector element in the equivariant feature, $\mathbf{f}_\text{eqv}(g)\in\bR^C$, is computed via rotating $\mathbf{f}$ by $g^{-1}\in G$ before passing it into a computational layer.
One can verify that this rotate-and-compute operation indeed generates $\mathbf{f}_\text{eqv}$ that is equivariant to any $g\in G$. Intuitively this means that, if an input is rotated by some $g_0\in G$, its equivariant feature $\mathbf{f}_\text{eqv}$ will undergo a \emph{circular shift} without changing any values, becoming $\mathbf{f}'_\text{eqv}$ such that $\mathbf{f}'_\text{eqv}(g') = \mathbf{f}_\text{eqv}(g)$, where $g'$ and $g$ satisfy $g'=g_0\circ g$. In other words, rotating the input $x\in \mathbb{R}^3$ by $g_0$ leads to a ``rotation'' by $g_0$ defined on the equivariant feature (in this case, the ``rotation'' is defined as shifting). For formal expressions and rigorous derivations, we refer readers to Chen~\etal~\cite{chen2021equivariant}.\footnote{We actually adopt a generalized version of EPN for any translation-equivariant backbone networks including convolutional backbones such as KPConv~\cite{thomas2019kpconv} and MLP backbones such as PointNet++~\cite{qi2017pointnet++}. Although EPN formulation is derived from continuous convolution, its discretized implementation admits an equivalent formulation. That is, rotating filters is equivalent to inversely rotating the input point cloud.}

Such equivariance network extracts very expressive equivariant features regarding input rotations. However, since the equivariance design is regarding the entire input point cloud without spatial scale concepts, EPN is not able to handle object-level equivariance when consuming a 3D scene.

Inspired by EPN, we achieve rotation equivariance via this feature augmenting strategy (sometimes referred to as feature orbits or filter orbits). Yet in contrast to EPN, our method allows learning \emph{object-level} equivariant features in full scene-scale inputs for detection.


\section{\modelfull}

\begin{figure*}[t]
    \centering
    \includegraphics[width=0.98\linewidth]{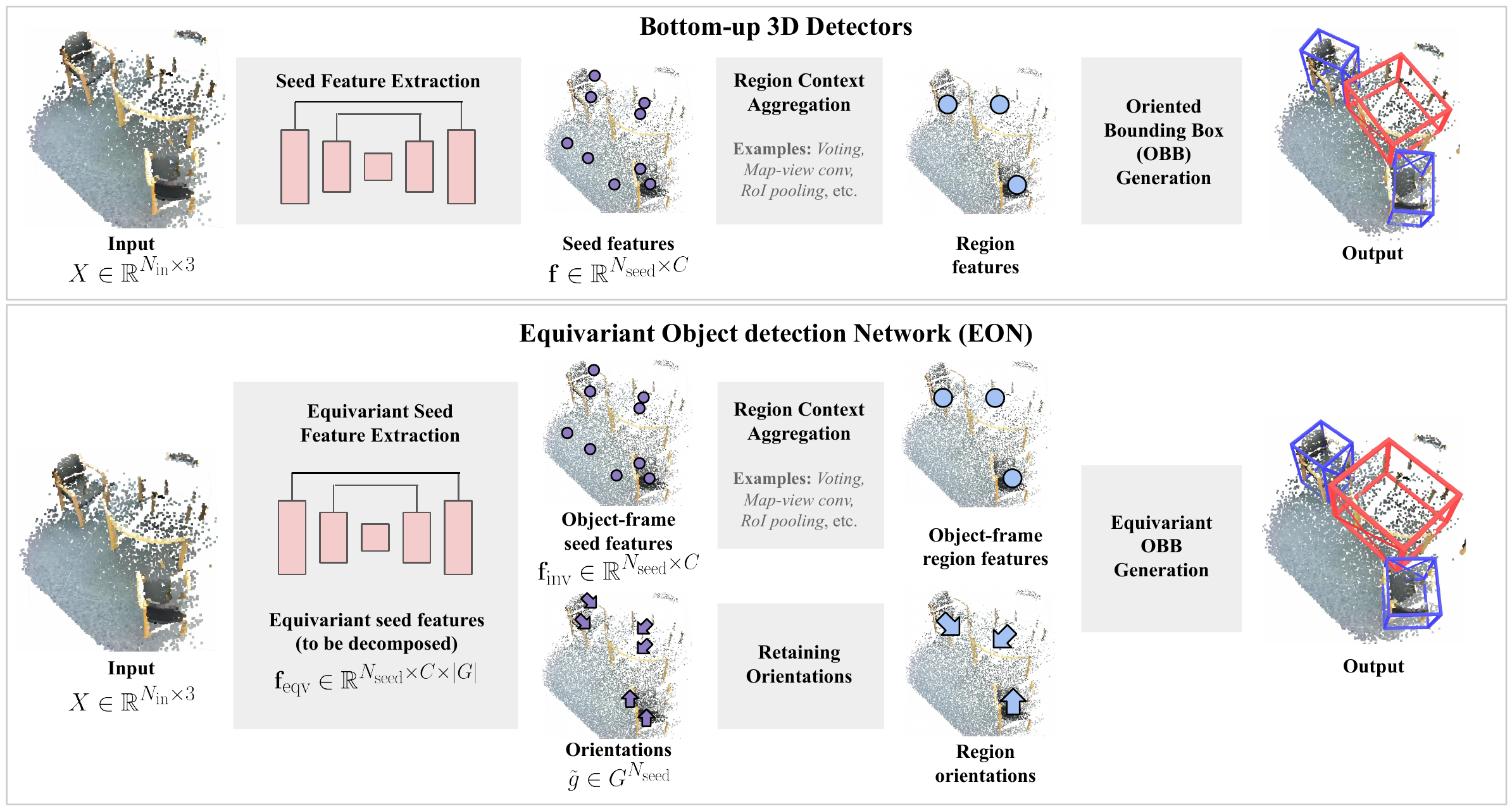}
    \vspace{-0.3cm}
    \caption{Overview of our technical approach. \textbf{Top}: We ground our method on popular bottom-up detector designs, including three modules (i.e., seed feature extraction, region aggregation, and OBB generation). \textbf{Bottom}: (\textit{Equivariant seed feature extraction}) In our modular design, we maintain equivariance computation up to seed features which are expected to capture object-level information. (\textit{Decomposed region context aggregation}) To maintain object-level equivariance, we decompose equivariant seed features into orientation hypotheses and object-frame invariant features. We aggregate the invariant geometric features, but not orientations (they are ``suspended''). (\textit{Equivariant OBB generation}) We generate object-frame OBBs from the region features and resume their orientation hypotheses to rotate them back to the scene frame.}
    \vspace{-0.2cm}
    \label{fig:idea_overview}
\end{figure*}

Our goal is to explicitly leverge object rotation equivariance in 3D detectors, to increase the network expressivity and discriminative power. We target at a modular design that allows us to directly equip state-of-the-art 3D detection models with object-level rotation equivariance.

We ground our method on popular bottom-up detector designs, which have been adopted by many state-of-the-art point cloud object detectors,
such as VoteNet~\cite{qi2019deep}, PointPillar~\cite{lang2019pointpillars}, PointRCNN~\cite{shi2019pointrcnn} and CenterPoint~\cite{yin2021center}. These methods generally include three modules: \textbf{seed feature extraction} in which a backbone processes point clouds into a dense set of features, \textbf{region context aggregation} that summarizes spatial regions (such as vote clustering~\cite{qi2019deep}, map-view convolution~\cite{yin2021center,lang2019pointpillars}, and RoI pooling~\cite{shi2019pointrcnn}) to produce a sparse set of features, and \textbf{OBB (oriented bounding box) generation} that proposes candidates from each regional feature. We show an illustration on the top of Figure \ref{fig:idea_overview}. We adapt these modules for object-level equivariance. 

There are two key challenges in our designs. The first is how to determine a proper local spatial support for equivariant feature computation so that the rotation equivariance is restricted to the object level. The second is how to aggregate contexts without breaking object-level equivariance. We propose \modelfull (\model), which incorporates our \textbf{rotation equivariance suspension} mechanism to address both challenges. We show an illustration of \model in the bottom of Figure \ref{fig:idea_overview}. 


\subsection{Overview}

Our main idea is that we let our model compute equivariant features up to an intermediate stage that is expected to be associated with object-level features. After this intermediate stage, we aggregate context geometric information only, as context orientations can break object-level rotation equivariance. To do this, we decompose an equivariant feature into an orientation hypothesis and an invariant feature in the object frame. The following context aggregation will only happen on the invariant geometric features (i.e., equivariance ``suspension''). Then, we generate OBB proposals from the aggregated invariant features, so the OBB proposals are also represented in the local object frame instead of the global scene frame. Finally, we rotate the object-frame proposals back to scene-frame OBBs using the object feature orientation (i.e., ``resuming'' the object orientation). See the bottom of Figure \ref{fig:idea_overview} for illustration.

One can conceptually verify the object-level equivariance in an ideal case.
When an object rotates in a static scene, the associated object-level feature orientation changes in the same way, while the object-frame feature remains unchanged. Therefore, the object-frame OBB proposals are invariant to the rotation, and hence the scene-frame OBBs do not change their shapes while rotating equivariantly with the objects. Notice that such equivariant design not only facilitates the equivariant detection outcomes, but more importantly allows learning better geometric object features~\cite{deng2021vector,cohen2016group,chen2021equivariant}. We ground our idea onto the bottom-up detector design (Figure \ref{fig:idea_overview}). 

\myparagraph{Seed feature extraction.} \model extracts equivariant seed features and suspends equivariance before the region aggregation stage where large-scale context information gets exchanged. Since this suspension happens in an intermediate stage, the deep detection network can adaptively adjust the effective spatial support~\cite{luo2016understanding}.

\myparagraph{Region context aggregation.} To model contexts, \model decomposes each equivariant seed feature by predicting an orientation and extracting an object-frame seed feature.
Ideally, object-frame seed features should be invariant to object poses and provide only geometric information for context aggregation.
Therefore, only object-frame seed features are fed into the region aggregation module to capture useful context information such as co-occurrences of objects.

\myparagraph{OBB generation.} Finally, the object-level equivariant OBB generation module \emph{resumes} the orientation information. It generates proposals in the object frame and transforms the proposals back to the scene frame using the predicted orientations. In the following, we describe \model's three modules in detail and how to plug them into modern 3D detectors.

\subsection{Equivariant seed feature extraction}\label{sec:method:seed}

\model extracts equivariant seed features and expects them to capture object-level information. In the following we focus on a single point output of a single layer, in order to make the symbols concise. Extending them to full point clouds and to entire networks is straightforward.

Given a layer $\psi: \bR^{N\times C'}\rightarrow \bR^C$ that maps some input point features $X\in\bR^{N\times C'}$ to a point feature $\mathbf{f}\in\bR^C$ (such as a KPConv~\cite{thomas2019kpconv} or a PointNet~\cite{qi2017pointnet}), its equivariant computation to map $X_\text{eqv}\in\bR^{N\times C'\times|G|}$ to $\mathbf{f}_\text{eqv}=\psi(X_\text{eqv})\in\bR^{C
\times|G|}$ is defined by:
\begin{align}
    \mathbf{f}_\text{eqv}(g) = \psi(X_\text{eqv}(g))\in\bR^C, g\in G,
\end{align}
where $G$ is a discrete subgroup of SO(3). For example, if one only considers rotation along a single axis, i.e., SO(2), a discretization can be $\{0, \pi/2, \pi, 3\pi/2\}$. And $X_\text{eqv}\in\bR^{N\times C'\times|G|}$ denotes the input equivariant features and the number of input points $N$ can vary. For the first layer where the input $X\in\bR^{N\times C'}$ is non-equivariant points, we pre-augment it to $X_\text{eqv}$ such that $X_\text{eqv}(g)=T_g^{-1}[X]$, with $T_g^{-1}$ denoting the inverse rotation of $g$. We follow EPN~\cite{chen2021equivariant} to attach a 1D convolution over the $|G|$ channel for each layer $\psi$ to further increase expressivity. Our formulation allows a drop-in replacement for most detection backbones. This equivariant computation is performed for the backbone seed feature extraction network.

\myparagraph{Objectness-aware equivariant feature decomposition.} We expect the equivariant seed feature extraction module to capture object-level geometric features. Thus, we suspend equivariant feature computation after this module and let the deep network to adaptively learn a proper effective receptive field~\cite{luo2016understanding}. To suspend equivariance, we decompose each equivariant feature $\mathbf{f}_\text{eqv}$ into an invariant feature represented in an object frame $\mathbf{f}_\text{inv}$, as well as the orientation $\tilde{g}$ of the object frame w.r.t. the scene frame. However, although object frame and orientation are well defined for foreground objects, these concepts do not hold for background stuff such as wall and road. Thus, we propose objectness-aware feature orbit decomposition which distinguishes between foreground and background points.

Specifically, we define foreground points as those inside an OBB, and all other points as background points. To predict the objectness for a seed point, we may jointly train a segmentation head attached to the seed features.
Then, our objectness-aware feature decomposition produces object-frame features as:
\begin{align}
    \mathbf{f}_{\text{inv}} =
    \begin{cases}
        \mathbf{f}_\text{eqv}(\tilde{g})\in\bR^C, \quad &\text{if it is foreground}\\
        \text{maxpool}(\{\mathbf{f}_\text{eqv}(g)\}_{g\in G})\in\bR^C, \quad &\text{otherwise}
    \end{cases}
\label{eqn:object_frame_feature}
\end{align}
where $\tilde{g}$ denotes a predicted orientation given by an orientation classification head $H$,
$\tilde{g} = \arg\max_g H(\mathbf{f}_\text{eqv}(g))$,
and $\text{maxpool}$ denotes max pooling over all $g\in G$. Notice we abuse the term ``object-frame seed feature'' to also cover background points.
The prediction head $H$ is jointly trained using a softmax classification loss. We generate the ground-truth orientation label by discretizing the orientation angle of each ground-truth OBB into $|G|$ bins, and then assigning the bin labels to all points inside the OBB. For background points, the orientations are undefined, and we ignore them in the rest of a forward pass.




\subsection{Decomposed region context aggregation}

In state-of-the-art 3D detectors, the context aggregation module summarizes features within a large spatial volume, allowing the following OBB generation to extract useful contextual information. Examples include the vote clustering in VoteNet~\cite{qi2019deep}, RoI pooling in PointRCNN~\cite{shi2019pointrcnn}, and Map-view convolution in CenterPoint~\cite{yin2021center} and PointPillar~\cite{lang2019pointpillars}. 

In our region context aggregation module, we aim to not only aggregate contextual geometric information, but also maintain object-level equivariance. Thus, we have two technical goals. The first goal is to aggregate invariant seed features for a sparse set of regions. We use the original region aggregation module from the detector. For every region, it takes as input the seed features inside the region, and outputs an object-frame invariant region feature.

The second goal is to retain the orientation of an object-of-interest for the region, and filter out context orientations. 
Ideally, the object-of-interest in a region is defined according to the proposal label assigned to the central point of the region, such as the target assignment methods used in PointRCNN~\cite{shi2019pointrcnn} and CenterPoint~\cite{yin2021center}. In this case, the output object-of-interest orientation $\tilde{h}$ is simply the prediction $\tilde{g}$ at the central point of the region.

However, for detectors that use IoU threshold-based target assignment, the object-of-interest for a region is not naturally defined, because the proposals from a region are dynamically associated with nearby ground-truth OBBs. 
In this case, we take the mode orientation in the region.
The idea is that if most foreground points in the region are from the same object, it is likely that the proposals are also mostly for that object. Thus we can capture its orientation by a mode selection. For regions that do not have a significant mode, the proposals are likely to be low-quality and get filtered out, or simply not be associated to any targets. Thus, they would yield little negative effects on the outcomes.

\subsection{Equivariant OBB generation}

Given an object-frame region feature and its orientation, our object-level equivariant OBB generation module yields OBBs in the scene frame. This is divided into two steps. First, we generate OBBs in the object frame by using the original module. Then, we transform the OBBs back to the scene frame using the region orientation. Specifically, given the center (or corner points) $c_{\text{inv}}\in \bR^3$ and the orientation $\theta_{\text{inv}}$ of an object-frame OBB, we transform the box to scene frame by replacing the orientation with $\theta=\theta_{\text{inv}} - \tilde{h}$ and replacing the center with $c=R^{-1}c_{\text{inv}}$ where $R$ is the rotation matrix for $\tilde{h}$.
\section{Experiments}

In this section we apply our \model design to modern detection models with various backbones and we show experimental results on several indoor and outdoor benchmarks.

\subsection{Experiment setup}
\myparagraph{Datasets.} We adopt ScanNetV2 dataset \cite{dai2017scannet} and SUN RGB-D dataset \cite{song2015sun} for indoor 3D detection. ScanNetV2 provides 1513 indoor scans with semantic segmentation labels. For benchmarking oriented object detection, we use the detection labels from Scan2CAD \cite{avetisyan2019scan2cad} which provides CAD models with oriented bounding boxes aligned with common objects in the scans. We report performances on 9 categories which have more than 200 instances and put all others into the ``Others'' class. SUN RGB-D dataset has around 5K RGB-D images annotated with amodal oriented boxes for 37 object classes. We follow the same setup as VoteNet \cite{qi2019deep} and report performances on 10 classes for SUN RGB-D. For outdoor 3D detection, we use KITTI \cite{geiger2012we} for evaluation, which contains 7481 training samples and 7518 testing samples. We follow the original evaluation protocol in KITTI.

\myparagraph{Implementation details.}
We implement our \model design for three 3D detectors including VoteNet \cite{qi2019deep} and a transformer-based state-of-the-art method from Liu et.al.~\cite{liu2021} for indoor scenes, and PointRCNN \cite{shi2019pointrcnn} for outdoor scenes. We plug in our proposed modules to replace their original modules, and denote the resultant models as \model-VoteNet, \model-Liu et.al., and \model-PointRCNN. As for the seed feature extraction module, all three methods originally use PointNet++ \cite{qi2017pointnet++} as their backbones. To demonstrate the applicability of our \model to different backbones, we use KPConv \cite{thomas2019kpconv} as the backbone for VoteNet. We show the KPConv backbone architecture in the supplement. Then we replace the original seed feature extraction modules as described in Sec.~\ref{sec:method:seed}. To predict orientation for each equivariant seed feature, we use a two-layer MLP for the head $H_o$. Since most objects are subject to gravity constraints, we only consider one degree of freedom (i.e., yaw) in the rotation group. We discretize it into 4 bins for classification. To predict the foreground segmentation, we use another two-layer MLP head for binary classification.

The region aggregation module of VoteNet consists of a voting stage which generates a spatial translation (``vote'') for each seed point, a grouping stage using the updated spatial coordinates, and a summarization stage to generate region features. Since the grouping happens in the scene frame, we inversely rotate the vote to the scene frame for the grouping. The region orientation $\tilde{h}$ is determined by mode selection. PointRCNN's region aggregation module is an RoI pooling where the object-of-interest is well defined. Its orientation $\tilde{h}$ uses the seed point that generates the RoI.

For all other settings such as the detection heads, input resolution, preprocessing, hyper-parameters and training configurations, we follow the implementations provided by the authors\footnote{VoteNet: \url{github.com/facebookresearch/votenet}, commit: 2f6d6d3. Liu et.al.: \url{https://github.com/zeliu98/Group-Free-3D}, commit: ef8b7bb. PointRCNN: \url{github.com/open-mmlab/OpenPCDet}, version: 0.3.0, commit: c9d31d3.}. We also use the same settings for our \model-VoteNet and \model-PointRCNN as their vanilla counterparts.

\begin{figure*}[t]
    \centering
    \includegraphics[width=0.85\linewidth]{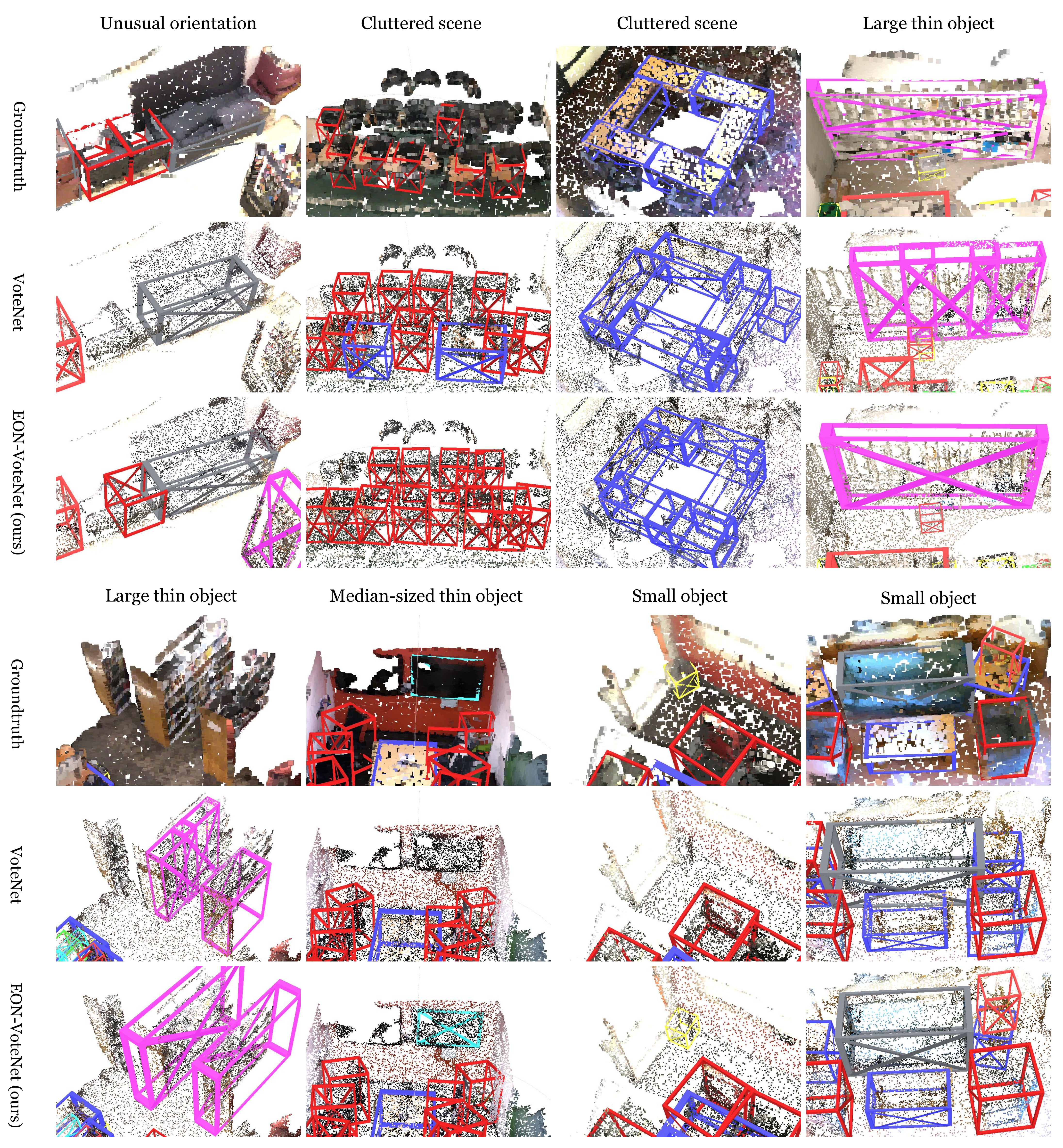}
    \vspace{-0.3cm}
    \caption{Qualitative results on ScanNetV2 dataset. Each row depicts a typical case where the object-level equivariance design shows promising improvements. Color information is not used for models, but only for visualization.}
    \vspace{-0.3cm}
    \label{fig:qualitative}
\end{figure*}

\subsection{Indoor datasets}

For 3D detection on indoor scenes, we compare our \model-VoteNet to VoteNet on the ScanNetV2 and the SUN RGB-D datasets. 

\myparagraph{ScanNetV2.}
For ScanNetV2 dataset, we show comparison results on validation set in Table~\ref{tbl:cmp_scan2cad}. As we can see, \model outperforms the vanilla methods significantly, with a boost in mAP by 6.3\% for VoteNet and 9.0\% for the transformer-based Liu et.al.~\cite{liu2021}. Notably, the performance gains are particularly significant for objects that are ``thin'' from the top view, such as display (+13.0\%/+37.0\% AP for VoteNet/Liu et.al.). We suppose a major reason be that for these objects, the accurate orientation prediction is crucial, and thus object-level equivariance design kicks in to help. Significant boost is also obtained for small objects like trash-bin (+17.9\%/7.4\% AP), which might be due to more accurate vote orientations brought by object-level equivariance, leading to better pinning-down of small objects' center positions.

\begin{table*}[t]
    \centering
    \small
    \begin{tabular}{lccccccccccc}
        \toprule
        Method & Trash-bin & Display & Others & Bathtub & Chair & Cabinet & Bookshelf & Table & Sofa & Bed & mAP \\
        \midrule
        VoteNet~\cite{qi2019deep} & 27.4&22.8&11.6&40.1&84.1&47.5&45.3&72.3&67.4&85.4&50.4 \\
        \ourvn (ours) & \textbf{45.3}&\textbf{35.8}&\textbf{16.4}&\textbf{49.1}&\textbf{86.3}&\textbf{51.9}&\textbf{51.0}&\textbf{75.0}&\textbf{68.7}&\textbf{87.2}&\textbf{56.7} \\
        \midrule
        Liu et.al.~\cite{liu2021} & 51.1 & 16.5 & 24.2 & \textbf{38.8} & 88.0 & 49.2 & 33.1 & 69.7 & 63.7 & 89.7 & 52.4\\
        \model-Liu et.al. (ours)  & \textbf{63.5} & \textbf{27.6} & \textbf{33.4} & 38.3 & \textbf{89.9} & \textbf{50.6} & \textbf{35.6} & \textbf{71.0} & \textbf{66.1} & \textbf{96.0} & \textbf{57.2}\\
        \bottomrule
    \end{tabular}
    \vspace{-0.3cm}
    \caption{Performance (AP$_\text{25}$) comparison on ScanNet V2 validation set using Scan2CAD detection labels.}
    \label{tbl:cmp_scan2cad}
    \vspace{-0.4cm}
\end{table*}

\myparagraph{SUN RGB-D.}
We also validate the effectiveness of our \model on SUN RGB-D dataset, in which the single-view derived point clouds are highly incomplete compared to the scans in ScanNetV2. SUN RGB-D is more difficult as the partial point clouds pose challenges not only on detection itself, but also on orientation estimation. We show comparison results on SUN RGB-D dataset in Table \ref{tbl:cmp_sunrgbd}. We draw the same observation as on ScanNetV2 that our \model-VoteNet outperforms VoteNet on all categories, leading to a boost in mAP by 3.1\%, despite the difficulty in orientation prediction. Similarly, the performance gain is more significant for thin object categories such as bookshelf (+8.9\% AP) and bathtub (+10.2\% AP).

\begin{table*}[t]
    \centering
    \small
    \begin{tabular}{lccccccccccc}
        \toprule
        Method & Nightstand & Toilet & Chair & Bathtub & Dresser & Desk & Table & Bookshelf & Sofa & Bed & mAP \\
        \midrule
        VoteNet~\cite{qi2019deep} & 58.4&89.9&74.5&68.8&34.6&26.1&48.5&22.3&65.9&84.9&57.4 \\
        \ourvn (ours) &\textbf{60.0}&\textbf{91.6}&\textbf{75.9}&\textbf{79.0}&\textbf{35.4}&\textbf{27.2}&\textbf{49.6}&\textbf{31.2}&\textbf{68.1}&\textbf{86.7}&\textbf{60.5} \\
        \bottomrule
    \end{tabular}
    \vspace{-0.2cm}
    \caption{Comparison of VoteNet with and without our \model design on SUN RGB-D dataset. Performances are measure by AP$_\text{25}$.}
    \label{tbl:cmp_sunrgbd}
    \vspace{-0.3cm}
\end{table*}

\subsection{Outdoor dataset}

For outdoor scenes, we evaluate our \model on KITTI 3D object detection dataset by equipping PointRCNN with our module designs, denoted as \model-PointRCNN. We show the comparison on the KITTI validation set in Table~\ref{tbl:cmp_kitti}. We observe that our \model-PointRCNN outperforms the vanilla model at all difficulty levels for Car and Pedestrian. The performance boost is most significant on the pedestrian category, where the AP gains are 3.02\%, 5.15\%, 2.90\% for Easy, Moderate, and Hard subsets, respectively. We conjecture that this might be because the object-level rotation equivariance is more useful when object orientations have a large variation, which is the case for indoor objects and pedestrians in outdoor scenes. Thus, our \model design has significant performance gains on indoor scenes and pedestrians on KITTI. As for cars and cyclists, they are mostly axis-aligned with less orientation variations. Thus, the performance gains are less significant.

Notice that \model-PoinRCNN and \model-VoteNet adopt different backbone networks (PointNet++ and KPConv, respectively). The consistent improvements across indoor/outdoor scenes and different models validate the wide applicability of our \model as a modular design.

\begin{table*}[t]
    \centering
    \small
    \begin{tabular}{lccccccccc}
        \toprule
        \multirow{2}{*}{Method} & \multicolumn{3}{c}{Car (IoU=0.7)} & \multicolumn{3}{c}{Pedestrian (IoU=0.5)} & \multicolumn{3}{c}{Cyclist (IoU=0.5)} \\
        \cmidrule(lr){2-4}\cmidrule(lr){5-7}\cmidrule(lr){8-10}
        &Easy & Moderate & Hard & Easy & Moderate & Hard & Easy & Moderate & Hard  \\
        \midrule
        PointRCNN~\cite{shi2019pointrcnn} &88.39&78.29&77.47&64.12&55.93&51.35&\textbf{87.81}&72.71&67.25  \\
        \model-PointRCNN (ours) &\textbf{89.11}&\textbf{78.61}&\textbf{77.55}&\textbf{67.14}&\textbf{61.08}&\textbf{54.25}&87.33&\textbf{73.36}&\textbf{67.41} \\
        \bottomrule
    \end{tabular}
    \vspace{-0.2cm}
    \caption{Results on KITTI 3D detection validation set. IoU threshold for AP is 0.7 for Car and 0.5 for Pedestrian/Cyclist, respectively.}
    \vspace{-0.5cm}
    \label{tbl:cmp_kitti}
\end{table*}

\subsection{Analysis}

To provide some insights on how our \model helps in 3D detection, we conduct several analysis experiments on ScanNetV2 dataset. We show further analysis on using category-level pose estimation, orientation discretization, suspension versus pooling, object rotation augmentation, and time and numbers of parameters in the supplementary material.

\myparagraph{Where to suspend equivariance suspension.}
Our core design in \model is suspending equivariance at the region aggregation stage to extract object-level equivariant features. We evaluate our design on VoteNet by suspending equivariance at earlier or latter stages, including suspending at the second last backbone layer (denoted as ``Pre\model'') and at the proposal generation stage (``Full\model''). ``Full\model'' keeps equivariant computation across the whole feature extraction, thus it is using full scene-level equivariance. We show the results in Table \ref{tbl:abl_full_eqv}. 

We draw two major observations. First, our \model compares much favorably with using full scene-level equivariance. This shows the rotation equivariance to full visual content does not help much in object detection, as the target object can rotate arbitrarily when the rest of the scene context remains unchanged. Second, our design to suspend equivariant computation at region aggregation yields best results, supporting our assumption that input to region aggregation roughly corresponds to an object level.

\begin{table}[t]
    \centering
    \small
    \begin{tabular}{lcc}
        \toprule
        Method & mAP@0.25 & mAP@0.5 \\
        \midrule
        Pre\model-VoteNet &54.5 &34.1 \\
        Full\model-VoteNet &52.7 &26.9 \\
        \ourvn (ours) & \textbf{56.7} & \textbf{36.5} \\
        \bottomrule
    \end{tabular}
    \vspace{-0.2cm}
    \caption{Evaluation on where to suspend equivariance on ScanNetV2 dataset with Scan2CAD detection labels.}
    \vspace{-0.4cm}
    \label{tbl:abl_full_eqv}
\end{table}

\myparagraph{Qualitative comparisons.}
We show qualitative examples in Figure \ref{fig:qualitative} to depict several typical cases where our \model shows great effectiveness.

In the first column, first row of Figure \ref{fig:qualitative}, we show that our \model-VoteNet is able to detect objects with unusual orientations. In the shown example, the black chairs are facing the wall, which is rare. Without having object-level equivariance by design, VoteNet misses these chairs. Nevertheless, equipped with \model, it successfully detects a chair with a very unusual facing. Notice that \model-VoteNet is able to detect a bookshelf that is also in a less common orientation.

In the second columns we show results on a cluttered scene where many similar chairs are packed together with minor orientation differences among them. While the baseline VoteNet can detect most of the chairs, their shapes are distorted. This might be partially due to the different orientations among the chair instances. In contrast, our \model-VoteNet produces a nearly identical shape across the oriented chairs, showing the promise to explicitly handle object rotation symmetry. In the third column, we show another cluttered scene with 6 similar desks next to each other. Again, our \model-VoteNet is able to detect nearly equal-shaped desks, while VoteNet yields shape-varying boxes.

In the fourth column and the first two columns in the second row, we show the detection of thin objects (including displays and bookshelves) where \model-VoteNet improves VoteNet most significantly. This aligns with our quantitative observations in Tables~\ref{tbl:cmp_scan2cad} and \ref{tbl:cmp_sunrgbd}. 
In the last two columns in the second row, we show that our \model-VoteNet is robust to detect small objects such as trash-bins.

\myparagraph{Oracle case exploration.}
To explore the extent to which our \model can potentially bring improvements to the baseline methods, we show an oracle case exploration in Table \ref{tbl:abl_oracle}. In the oracle models, we use the ground-truth orientation labels (discretized into 4 bins) and foreground segmentation in both training and testing. This oracle model shows a strong improvement of +19.4\%/+18.7\% for AP$_\text{25}$ and AP$_\text{50}$ over the baseline VoteNet. When using only ground-truth orientation labels or segmentation, the performance boosts are also promising. Thus, we might expect further improvements along with better orientation prediction and segmentation prediction, possibly from external domain-tailored models.

\begin{table}[t]
    \centering
    \small
    \begin{tabular}{lcc}
        \toprule
        Method & mAP@0.25 & mAP@0.5 \\
        \midrule
        VoteNet~\cite{qi2019deep} &50.4 &28.3 \\
        \ourvn (ours) & \textbf{56.7} & \textbf{36.5} \\
        \midrule
        Oracle (ori.) &64.0 & 42.9 \\
        Oracle (seg.) &65.3&40.6\\
        Oracle (ori. and seg.) & \textbf{69.8} & \textbf{47.0} \\
        \bottomrule
    \end{tabular}
    \vspace{-0.2cm}
    \caption{Comparison to oracle model on ScanNetV2 dataset with Scan2CAD detection labels.}
    \vspace{-0.4cm}
    \label{tbl:abl_oracle}
\end{table}

\section{Conclusion}
In this work we explore how to leverage object-level rotation equivariance for 3D object detection. To this end, we propose \modelfull (\model) with our core design to suspend rotation equivariance at an intermediate stage in the feature learning backbone. We apply our design to various backbones and models for both indoor and outdoor scenes. Our experiments show that \model can consistently improve the state-of-the-art detectors, indicating the effectiveness of explicitly modeling object-level equivariance in detection models.

\myparagraph{Limitations.}
One major limitation is the jointly learned orientation/segmentation predictions which might be less accurate than being treated separately. As suggested by the oracle experiments, while our current implementation can already benefit from the object-level equivariance, it is far from reaching the potential maximum gains. Nevertheless, our results suggest the promise of leveraging object-level equivariance for 3D detection. Another limitation is that we assume 1D rotation along the gravity axis. While extending to 3D rotation is straightforward, further exploring it might be beneficial for scenarios such as robotic manipulation.

\myparagraph{Acknowledgements.} This work is in part supported by the Qualcomm Innovation Fellowship, the Stanford HAI, Center for Integrated Facility Engineering, Toyota Research Institute (TRI), Samsung, Autodesk, Amazon, and Meta. This work was partly done when Yu was visiting Shanghai Qi Zhi Institute. We thank Fr\'edo Durand for helpful discussions.

{\small
\bibliographystyle{ieee_fullname}
\bibliography{reference_ky}

\begin{thebibliography}{10}\itemsep=-1pt

\bibitem{avetisyan2019scan2cad}
Armen Avetisyan, Manuel Dahnert, Angela Dai, Manolis Savva, Angel~X Chang, and
  Matthias Nie{\ss}ner.
\newblock Scan2cad: Learning cad model alignment in rgb-d scans.
\newblock In {\em CVPR}, 2019.

\bibitem{chen2021equivariant}
Haiwei Chen, Shichen Liu, Weikai Chen, Hao Li, and Randall Hill.
\newblock Equivariant point network for 3d point cloud analysis.
\newblock In {\em CVPR}, 2021.

\bibitem{cohen2019gauge}
Taco Cohen, Maurice Weiler, Berkay Kicanaoglu, and Max Welling.
\newblock Gauge equivariant convolutional networks and the icosahedral cnn.
\newblock In {\em ICML}, 2019.

\bibitem{cohen2016group}
Taco Cohen and Max Welling.
\newblock Group equivariant convolutional networks.
\newblock In {\em ICML}, 2016.

\bibitem{cohen2018spherical}
Taco~S Cohen, Mario Geiger, Jonas K{\"o}hler, and Max Welling.
\newblock Spherical cnns.
\newblock {\em arXiv:1801.10130}, 2018.

\bibitem{dai2017scannet}
Angela Dai, Angel~X Chang, Manolis Savva, Maciej Halber, Thomas Funkhouser, and
  Matthias Nie{\ss}ner.
\newblock Scannet: Richly-annotated 3d reconstructions of indoor scenes.
\newblock In {\em CVPR}, 2017.

\bibitem{deng2021vector}
Congyue Deng, Or Litany, Yueqi Duan, Adrien Poulenard, Andrea Tagliasacchi, and
  Leonidas Guibas.
\newblock Vector neurons: A general framework for so(3)-equivariant networks.
\newblock {\em ICCV}, 2021.

\bibitem{esteves2018learning}
Carlos Esteves, Christine Allen-Blanchette, Ameesh Makadia, and Kostas
  Daniilidis.
\newblock Learning so(3) equivariant representations with spherical cnns.
\newblock In {\em ECCV}, 2018.

\bibitem{geiger2012we}
Andreas Geiger, Philip Lenz, and Raquel Urtasun.
\newblock Are we ready for autonomous driving? the kitti vision benchmark
  suite.
\newblock In {\em CVPR}, 2012.

\bibitem{han2021redet}
Jiaming Han, Jian Ding, Nan Xue, and Gui-Song Xia.
\newblock Redet: a rotation-equivariant detector for aerial object detection.
\newblock In {\em CVPR}, 2021.

\bibitem{lang2019pointpillars}
Alex~H Lang, Sourabh Vora, Holger Caesar, Lubing Zhou, Jiong Yang, and Oscar
  Beijbom.
\newblock Pointpillars: Fast encoders for object detection from point clouds.
\newblock In {\em CVPR}, 2019.

\bibitem{li2019discrete}
Jiaxin Li, Yingcai Bi, and Gim~Hee Lee.
\newblock Discrete rotation equivariance for point cloud recognition.
\newblock In {\em ICRA}, 2019.

\bibitem{li2021leveraging}
Xiaolong Li, Yijia Weng, Li Yi, Leonidas Guibas, A~Lynn Abbott, Shuran Song,
  and He Wang.
\newblock Leveraging se(3) equivariance for self-supervised category-level
  object pose estimation.
\newblock {\em arXiv:2111.00190}, 2021.

\bibitem{liu2021}
Ze Liu, Zheng Zhang, Yue Cao, Han Hu, and Xin Tong.
\newblock Group-free 3d object detection via transformers.
\newblock In {\em ICCV}, 2021.

\bibitem{luo2016understanding}
Wenjie Luo, Yujia Li, Raquel Urtasun, and Richard Zemel.
\newblock Understanding the effective receptive field in deep convolutional
  neural networks.
\newblock In {\em NeurIPS}, 2016.

\bibitem{pan20213d}
Xuran Pan, Zhuofan Xia, Shiji Song, Li~Erran Li, and Gao Huang.
\newblock 3d object detection with pointformer.
\newblock In {\em CVPR}, 2021.

\bibitem{qi2019deep}
Charles~R Qi, Or Litany, Kaiming He, and Leonidas~J Guibas.
\newblock Deep hough voting for 3d object detection in point clouds.
\newblock In {\em ICCV}, 2019.

\bibitem{qi2017pointnet}
Charles~R Qi, Hao Su, Kaichun Mo, and Leonidas~J Guibas.
\newblock Pointnet: Deep learning on point sets for 3d classification and
  segmentation.
\newblock In {\em CVPR}, 2017.

\bibitem{qi2017pointnet++}
Charles~R Qi, Li Yi, Hao Su, and Leonidas~J Guibas.
\newblock Pointnet++: Deep hierarchical feature learning on point sets in a
  metric space.
\newblock {\em arXiv:1706.02413}, 2017.

\bibitem{shi2020pv}
Shaoshuai Shi, Chaoxu Guo, Li Jiang, Zhe Wang, Jianping Shi, Xiaogang Wang, and
  Hongsheng Li.
\newblock Pv-rcnn: Point-voxel feature set abstraction for 3d object detection.
\newblock In {\em CVPR}, 2020.

\bibitem{shi2019pointrcnn}
Shaoshuai Shi, Xiaogang Wang, and Hongsheng Li.
\newblock Pointrcnn: 3d object proposal generation and detection from point
  cloud.
\newblock In {\em CVPR}, 2019.

\bibitem{shi2020point}
Weijing Shi and Raj Rajkumar.
\newblock Point-gnn: Graph neural network for 3d object detection in a point
  cloud.
\newblock In {\em CVPR}, 2020.

\bibitem{song2015sun}
Shuran Song, Samuel~P Lichtenberg, and Jianxiong Xiao.
\newblock Sun rgb-d: A rgb-d scene understanding benchmark suite.
\newblock In {\em CVPR}, 2015.

\bibitem{thomas2019kpconv}
Hugues Thomas, Charles~R Qi, Jean-Emmanuel Deschaud, Beatriz Marcotegui,
  Fran{\c{c}}ois Goulette, and Leonidas~J Guibas.
\newblock Kpconv: Flexible and deformable convolution for point clouds.
\newblock In {\em CVPR}, 2019.

\bibitem{thomas2018tensor}
Nathaniel Thomas, Tess Smidt, Steven Kearnes, Lusann Yang, Li Li, Kai Kohlhoff,
  and Patrick Riley.
\newblock Tensor field networks: Rotation-and translation-equivariant neural
  networks for 3d point clouds.
\newblock {\em arXiv:1802.08219}, 2018.

\bibitem{worrall2017harmonic}
Daniel~E Worrall, Stephan~J Garbin, Daniyar Turmukhambetov, and Gabriel~J
  Brostow.
\newblock Harmonic networks: Deep translation and rotation equivariance.
\newblock In {\em CVPR}, 2017.

\bibitem{xie2020mlcvnet}
Qian Xie, Yu-Kun Lai, Jing Wu, Zhoutao Wang, Yiming Zhang, Kai Xu, and Jun
  Wang.
\newblock Mlcvnet: Multi-level context votenet for 3d object detection.
\newblock In {\em CVPR}, 2020.

\bibitem{yang2021sampling}
Guo-Ye Yang, Xiang-Li Li, Ralph~R Martin, and Shi-Min Hu.
\newblock Sampling equivariant self-attention networks for object detection in
  aerial images.
\newblock {\em arXiv:2111.03420}, 2021.

\bibitem{yin2021center}
Tianwei Yin, Xingyi Zhou, and Philipp Krahenbuhl.
\newblock Center-based 3d object detection and tracking.
\newblock In {\em CVPR}, 2021.

\bibitem{zhang2020h3dnet}
Zaiwei Zhang, Bo Sun, Haitao Yang, and Qixing Huang.
\newblock H3dnet: 3d object detection using hybrid geometric primitives.
\newblock In {\em ECCV}, 2020.

\bibitem{zhou2018voxelnet}
Yin Zhou and Oncel Tuzel.
\newblock Voxelnet: End-to-end learning for point cloud based 3d object
  detection.
\newblock In {\em CVPR}, 2018.

\bibitem{zhu2019class}
Benjin Zhu, Zhengkai Jiang, Xiangxin Zhou, Zeming Li, and Gang Yu.
\newblock Class-balanced grouping and sampling for point cloud 3d object
  detection.
\newblock {\em arXiv:1908.09492}, 2019.

\end{thebibliography}
}

\end{document}


\title{Supplementary Material for \\
Rotationally Equivariant 3D Object Detection}

\author{Hong-Xing Yu \\
Stanford University\\
\and
Jiajun Wu\\
Stanford University\\
\and
Li Yi\\
Tsinghua University, Shanghai Qi Zhi Institute\\
}
\maketitle

\begin{abstract}
In this supplementary document, we provide further analysis in Section \ref{supp:sec:analysis}, including analysis on using category-level pose estimation, the number of discretized orientation bins, ablation study on rotation equivariance suspension, analysis on object rotation augmentations, and comparisons of inference time and the number of parameters. We also include additional implementation details of the backbone architecture of VoteNet in Section \ref{supp:sec:implementation}. Our code can be found in our supplementary material.
\end{abstract}
\section{Additional Experimental Analysis}\label{supp:sec:analysis}
In this section we provide additional experiments analysis for \model.

\myparagraph{Comparison to using category-level pose estimation.} A naive method to directly achieve object-level equivariance in 3D detection might be to use category-level pose estimation to refine the bounding box orientations. To do this, we attach a head to VoteNet to predict a NOCS~\cite{wang2019normalized} coordinate for each box proposal, and use the estimated pose to refine box orientation. Table \ref{tbl:nocs} shows that category-level pose estimation can only marginally improve the baseline VoteNet. This is because our equivariant design is not only to refine orientations. 
More importantly, it learns better geometric object features and distinguishes objects from backgrounds to improve proposal generations.

\begin{figure}[t]
    \centering
    \includegraphics[width=0.89\linewidth]{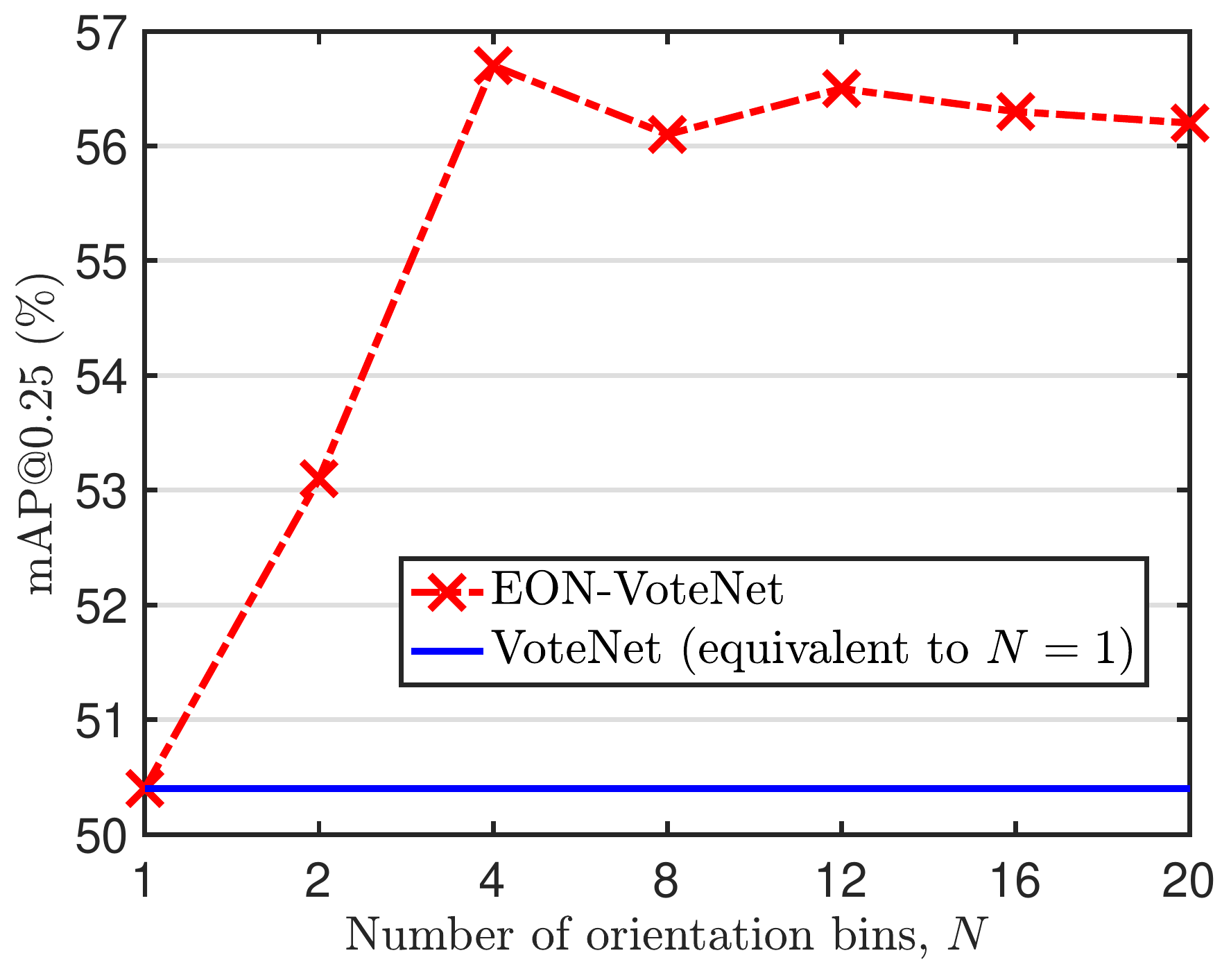}
    \vspace{-0.2cm}
    \caption{Performances versus different numbers of orientation bins. Results are evaluated on ScanNetV2 with Scan2CAD labels, measured by mAP@0.25. Note that \model-VoteNet is equivalent to VoteNet when $N=1$.}
    \label{fig:nrot}
    \vspace{-0.2cm}
\end{figure}

\myparagraph{Effects of finer orientation discretization.} In our \model we predict the object-level orientations by discretizing the yaw orientation group into $N$ bins and perform a 4-way classification for each seed feature orbit. To explore the effect of different numbers of bins, we study the mAP performance versus different values of $N$ on ScanNetV2 and show the results in Figure \ref{fig:nrot}. We can see that the performance boost of \model-VoteNet over the baseline VoteNet reaches the peak at $N=4$, and saturates with finer discretization. Thus, we empirically set $N=4$ to obtain full performance gain with minimal additional computation.

\begin{table}[t]
    \centering
    \small
    \begin{tabular}{lcc}
        \toprule
        Method & mAP@0.25 & mAP@0.5 \\
        \midrule
        VoteNet & 50.4 & 28.3 \\
        VoteNet+NOCS~\cite{wang2019normalized} &50.7 &28.5 \\
        \model-VoteNet (ours) & \textbf{56.7} & \textbf{36.5} \\
        \bottomrule
    \end{tabular}
    \vspace{-0.2cm}
    \caption{Comparison to category-level pose estimation method on ScanNetV2 dataset with Scan2CAD detection labels.}
    \vspace{-0.2cm}
    \label{tbl:nocs}
\end{table}
\begin{figure*}[t]
    \centering
    \includegraphics[width=0.8\linewidth]{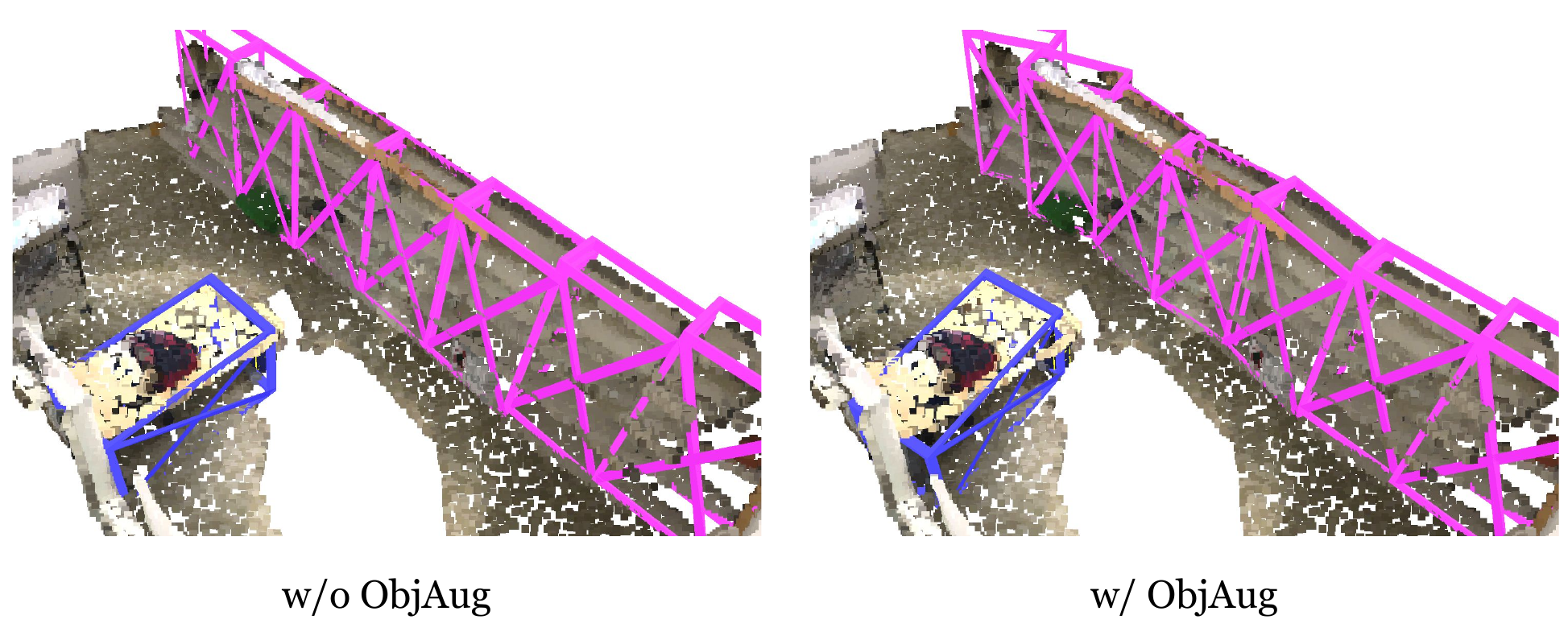}
    \vspace{-0.2cm}
    \caption{An example of using object rotation data augmentation.}
    \vspace{-0.2cm}
    \label{fig:obj_aug}
\end{figure*}
\begin{table*}[t]
    \centering
    \small
    \begin{tabular}{lccccccccccc}
        \toprule
        Method & Trash-bin & Display & Others & Bathtub & Chair & Cabinet & Bookshelf & Table & Sofa & Bed & mAP \\
        \midrule
        ION-VoteNet &34.0& 23.8&\textbf{20.0}&43.9&71.7&47.1&51.4&70.3&\textbf{69.9}&83.5&51.5 \\
        \ourvn (ours) & \textbf{45.3}&\textbf{35.8}&16.4&\textbf{49.1}&\textbf{86.3}&\textbf{51.9}&\textbf{51.0}&\textbf{75.0}&68.7&\textbf{87.2}&\textbf{56.7} \\
        \bottomrule
    \end{tabular}
    \caption{Ablation study on rotation equivariance suspension. We compare \model-VoteNet to ION-VoteNet on ScanNet V2 validation set using Scan2CAD detection labels. Performances are measure by AP$_\text{25}$.}
    \label{tbl:abl_eqv_baseline}
\end{table*}
\begin{table}[t]
    \centering
    \small
    \begin{tabular}{lcc}
        \toprule
        Method & VoteNet & \model-VoteNet \\
        \midrule
        Trained w/ ObjAug &49.7 & 53.6 \\
        Trained w/o ObjAug & \textbf{50.4} & \textbf{56.7} \\
        \bottomrule
    \end{tabular}
    \vspace{-0.2cm}
    \caption{Evaluation for object rotation data augmentation (ObjAug). Tested on ScanNetV2 with Scan2CAD labels, measured by mAP@0.25.}
    \vspace{-0.2cm}
    \label{tbl:obj_aug}
\end{table}
\begin{table}[t]
    \centering
    \small
    \begin{tabular}{lcc}
        \toprule
        Method & Inference time (ms) & \#Parameters \\
        \midrule
        VoteNet &85.3 &0.96M \\
        \model-VoteNet &92.8 &1.10M \\
        \bottomrule
    \end{tabular}
    \vspace{-0.2cm}
    \caption{Comparison on inference time (using a single input point cloud with 40K points as input) tested on a single TitanRTX, and numbers of parameters.}
    \label{tbl:time_and_param}
    \vspace{-0.2cm}
\end{table}
\myparagraph{Ablating rotation equivariance suspension.} A core component in our model design is the rotation equivariance suspension on the region aggregation module for object-level rotation equivariance. Thus, we explore the importance of rotation equivariance suspension by replacing it with an invariant max-pooling which is typically adopted for extracting invariant features from equivariant features \cite{deng2021vector,chen2021equivariant}. We refer to this baseline model as Invariant Object detection Networks, or ION. ION extracts the same feature orbits as \model in the seed feature extraction module. Instead of suspending the rotation equivariance, ION directly extracts invariant features from the feature orbits. We show a comparison of \model-VoteNet and ION-VoteNet in Table \ref{tbl:abl_eqv_baseline}.

From Table \ref{tbl:abl_eqv_baseline} we see that \model-VoteNet outperforms ION-VoteNet significantly, despite that the latter uses the same equivariant computation for seed features. This comparison suggests the importance of maintaining object-level equivariance throughout the detector.

\myparagraph{Effect of object rotation data augmentation.} 
A natural alternative idea to approach object rotation equivariance is to use object rotation data augmentation. That is, for each oriented box in a training scene, we randomly rotate the points inside the box w.r.t. its box central gravity axis by $[-30, 30]$ degrees. We also correspondingly rotate the box orientation label and vote labels inside the box. We denote this augmentation strategy as ``ObjAug''. We show results of VoteNet and \model-VoteNet when using ObjAug in Table \ref{tbl:obj_aug}.

From Table \ref{tbl:obj_aug} we can observe that ObjAug hurts both methods' performances. A possible reason is that the bounding boxes cannot perfectly enclose an object, so that they either include background segments or leave some object surface points unchanged (for example, some surface points of the desk is not rotated in Figure \ref{fig:obj_aug}). This leads to corruptions to the point cloud data. Another possible reason is that ObjAug changes the scene layouts to an unnatural configuration (for example, all bookshelves have different orientations as shown in Figure \ref{fig:obj_aug}), so that they deviate from common room layouts present in the test scenes.

\myparagraph{Inference time and numbers of parameters.} Finally, we report the inference time and the numbers of parameters in Table \ref{tbl:time_and_param} for VoteNet and \model-VoteNet. All models are tested on a single Nvidia TitanRTX with batch size 1. We can see that our \model does not add significant cost in inference time and model capacity.

\section{Additional Implementation Details}\label{supp:sec:implementation}
In our experiments, we apply our \model to VoteNet \cite{qi2019deep} and PointRCNN \cite{shi2019pointrcnn}. While both VoteNet and PointRCNN originally use PointNet++ \cite{qi2017pointnet++} as their backbones, we replace it with KPConv \cite{thomas2019kpconv} for VoteNet to test our \model design on different backbones. We show the detailed architecture of the KPConv-based backbone in Table \ref{tbl:backbone} for VoteNet and Table \ref{tbl:backbone_eqv} for \model-VoteNet.

\begin{table*}[t]
    \centering
    \setlength{\tabcolsep}{6pt}
    \renewcommand{\arraystretch}{1.2}
    \begin{tabular}{ccccc}
        
        \toprule
        \textbf{Layer name} & \textbf{Output shape} & \textbf{Radius (meter)}  &\textbf{Kernel dimension} & \textbf{Note} \\
        \midrule
        \texttt{SA1} & 2048$\times$32 & 0.2 & 15$\times$4$\times$32 &  \\
        \texttt{SA2} & 1024$\times$32 & 0.4 & 15$\times$32$\times$32 &  \\
        \texttt{SA3} & 512$\times$64 & 0.8 & 15$\times$32$\times$64 & Skip to \texttt{FP4} \\
        \texttt{SA4} & 256$\times$64 & 1.2 & 15$\times$64$\times$64 & Skip to \texttt{FP3} \\
        \texttt{SA5} & 128$\times$64 & 1.5 & 15$\times$64$\times$64 & Skip to \texttt{FP2} \\
        \texttt{SA6} & 64$\times$128 & 1.8 & 15$\times$64$\times$128 &  \\
        \texttt{FP1} & 128$\times$64 & 1.5 & 15$\times$192$\times$64 &  \\
        \texttt{FP2} & 256$\times$64 & 1.2 & 15$\times$128$\times$64 &  \\
        \texttt{FP3} & 512$\times$64 & 0.8 & 15$\times$128$\times$64 &  \\
        \texttt{FP4} & 1024$\times$64 & 0.4 & 15$\times$96$\times$64 &  \\
        \bottomrule
    \end{tabular}
    \vspace{5pt}
    \caption{Architecture of KPConv-based \cite{thomas2019kpconv} backbone used for VoteNet \cite{qi2019deep}. Following PointNet++ \cite{qi2017pointnet++} layer definition, ``SA'' in the layer name stands for set abstraction, and ``FP'' stands for feature propagation. As in PointNet++ \cite{qi2017pointnet++}, SA layers use farthest point sampling for abstraction downsampling, FP layers use 3-NN for upsampling, and every layer uses ball queries for local grouping. The format for output shape is $\#\texttt{points}\times\texttt{output\_dimension}$. The format for kernel dimension is $\#\texttt{kernels}\times\texttt{input\_dimension}\times\texttt{output\_dimension}$. Each layer is followed by a batchnorm and a ReLU activation.}
    \label{tbl:backbone}
\end{table*}
\begin{table*}[t]
    \centering
    \setlength{\tabcolsep}{6pt}
    \renewcommand{\arraystretch}{1.2}
    \begin{tabular}{ccccc}
        
        \toprule
        \textbf{Layer name} & \textbf{Output shape} & \textbf{Radius (meter)}  &\textbf{Kernel dimension} & \textbf{Note} \\
        \midrule
        \texttt{SA1} & 2048$\times$32$\times$4 & 0.2 & 15$\times$4$\times$32 &  \\
        \texttt{SA2} & 1024$\times$32$\times$4 & 0.4 & 15$\times$32$\times$32 &  \\
        \texttt{SA3} & 512$\times$64$\times$4 & 0.8 & 15$\times$32$\times$64 & Skip to \texttt{FP4} \\
        \texttt{SA4} & 256$\times$64$\times$4 & 1.2 & 15$\times$64$\times$64 & Skip to \texttt{FP3} \\
        \texttt{SA5} & 128$\times$64$\times$4 & 1.5 & 15$\times$64$\times$64 & Skip to \texttt{FP2} \\
        \texttt{SA6} & 64$\times$128$\times$4 & 1.8 & 15$\times$64$\times$128 &  \\
        \texttt{FP1} & 128$\times$64$\times$4 & 1.5 & 15$\times$192$\times$64 &  \\
        \texttt{FP2} & 256$\times$64$\times$4 & 1.2 & 15$\times$128$\times$64 &  \\
        \texttt{FP3} & 512$\times$64$\times$4 & 0.8 & 15$\times$128$\times$64 &  \\
        \texttt{FP4} & 1024$\times$64$\times$4 & 0.4 & 15$\times$96$\times$64 &  \\
        \bottomrule
    \end{tabular}
    \vspace{5pt}
    \caption{Architecture of KPConv-based \cite{thomas2019kpconv} equivariant backbone used for \model-VoteNet. Following PointNet++ \cite{qi2017pointnet++} layer definition, ``SA'' in the layer name stands for set abstraction, and ``FP'' stands for feature propagation. As in PointNet++ \cite{qi2017pointnet++}, SA layers use farthest point sampling for abstraction downsampling, FP layers use 3-NN for upsampling, and every layer uses ball queries for local grouping. The format for output shape is $\#\texttt{points}\times\texttt{output\_dimension}\times\texttt{\#orientation}$. The format for kernel dimension is $\#\texttt{kernels}\times\texttt{input\_dimension}\times\texttt{output\_dimension}$. Each layer is followed by a batchnorm and a ReLU activation, and a 1D convolution with kernel size 3 on the orientation dimension for orientation channel communication \cite{chen2021equivariant}.}
    \label{tbl:backbone_eqv}
\end{table*}

{\small
\bibliographystyle{ieee_fullname}
\bibliography{reference_ky}
}